# Transformer Based Bengali Chatbot Using General Knowledge Dataset

Abu Kaisar Mohammad Masum[1], Sheikh Abujar[2], Sharmin Akter[1], Nushrat Jahan Ria[1] and Syed Akhter Hossain[3]
[1] *Daffodil International University,*
[2] *Independent University of Bangladesh,*
[3] *University of Liberal Arts Bangladesh,*
Dhaka, Bangladesh
{abu.cse, sharminakter.cse, nushrat15-9771}@diu.edu.bd
abujar@iub.edu.bd
akhter.hossain@ulab.edu.bd
{kaisar, abujar, sharmin, ria, akhter}@cilab.ai

*Abstract*—An AI chatbot provides an impressive response after learning from the trained dataset. In this decade, most of the research work demonstrates that deep neural models superior to any other model. RNN model regularly used for determining the sequence-related problem like a question and it answers. This approach acquainted with everyone as seq2seq learning. In a seq2seq model mechanism, it has encoder and decoder. The encoder embedded any input sequence, and the decoder embedded output sequence. For reinforcing the seq2seq model performance, attention mechanism added into the encoder and decoder. After that, the transformer model has introduced itself as a high-performance model with multiple attention mechanism for solving the sequence-related dilemma. This model reduces training time compared with RNN based model and also achieved state-of-the-art performance for sequence transduction. In this research, we applied the transformer model for Bengali general knowledge chatbot based on the Bengali general knowledge Question Answer (QA) dataset. It scores 85.0 BLEU on the applied QA data. To check the comparison of the transformer model performance, we trained the seq2seq model with attention on our dataset that scores 23.5 BLEU.

## I. INTRODUCTION

NMT model performs better than normal SMT model. Translating source to target text is solved by encoder and decoder in the NMT model. After adding the attention with NMT model it increases performance of encoder and decoder [1]. In the encoder, a fixed-length vector is used and the decoder generates the output. All text is maintaining the sequence during learning and predicting. For that RNN, LSTM and GRU are used for managing the learning process in encoder and decoder [12] [13]. Be that as it may, SMT doesn't perform well and act increasingly complex by utilizing progressively extra highlights. Specialists concentrated on sequence learning advancements because of this impediment of SMT. Notwithstanding, neural system approaches have been presented in MT fields in the late 20-th century [14].

A chatbot is an intelligent system that response with humans based on the natural language text input [15]. Today used of chatbot application is increased day by day. Many areas such as social media, banking, customer care, medical those are used the chatbot for customer satisfaction. Therefore, an automatic chatbot with intelligent response is required for those areas. Deep learning algorithms provide a better solution for making these types of an intelligent system. Also, the result gives outperformance compared [16] with human and evaluation metric.

A chatbot, otherwise called a conversational specialist, is a PC programming fit for taking a characteristic language input and giving a conversational yield progressively. This human-chatbot association is commonly helped out through a graphical UI dependent on human-PC cooperation standards [17]. There are two types of chatbot in recent existing work such as generation and retrieval-based approaches. Selecting the response text from the current work is the main idea of retrieval-based approaches and generating the answer form related context is the core part of the generation based chatbot. In this research we applied transformer model for making an intelligent chatbot for Bengali general knowledge dataset. Transformer provides better performance than other sequence transduction model such as seq2seq model. We also applied seq2seq model with dot attention mechanism. Transformer model score better BLEU than seq2seq with attention mechanism. And the response of each question is accurate than seq2seq model. In section IV our model evaluation and response result are given. Methodology description is given in the section III.

## II. RELATED WORK

In most factual ways to deal with machine interpretation, the fundamental units of interpretation are phrases that are made out of at least one words. There is some limitation of SMT. Neural machine translation concept comes to solve machine translation problems. Kalchbrenner et al. [4] introduce recurrent approach for continuous translation. In this method sequence of text is memorized by the recurrent network than translate the text sequence wise. For known word recurrent methods work properly for NMT. But increasing the performance for unknown word or sentences Cho et al. [2] acquaint encoder and decoder for recurrent sentence translation. The input text encoded by the encoder and the output is decoded by the decoder. Both are followed sequence wise text translation. Increasing the performance of basic encoder and decoder Bahdanau et al. [1] added attention with the learning process. A fixed length of text vector is used for encoder and decoder predict the output related to this fixed length vector text. After that Luong et al. [8] recommend another approach in attention mechanism that used a global technique for sequence that works in all source word.

A simple chatbot application has two part such as question by user and response of the intelligent system. Seq2Seq learning produce a good result for these types of problem. The input question is learned by the encoder and output generate by the decoder. Basic RNN and LSTM is used for manage the sequence [18]. But long seq2seq learning approach sometimes loss information. And response of corresponding answer is not related to the asking question. So, attention with context is added into the seq2seq learning for improve the response and reduce the sequence loss [19].

Then context provide the output word using the attention and send sequence into the decoder. On the off chance that the vector measurement is fixed for the encoder and, at that point, the sentence is longer, so the encoder needs to encode the entire wellspring of sentences in a context and this is making an issue for the sentence with long factor length. What's more, taking care of this issue through consideration instrument by monitoring the concealed conditions of source memory and afterwards initiating those to the significant ones utilizing the context.

### III. METHODOLOGY

This section we have discussed the methodology part of this research. We gave the details description of encoder-decoder of the transformer, dataset properties and processing, problem understanding, transformer model architecture for our general knowledge chatbot. Overall model parameters are set on the basis of our own dataset.

*A. Dataset properties*

For the research, we used Bengali general knowledge-based question answering dataset. The dataset contains three attributes. But making the chatbot model we used only two attributes such as questions and their answers. The question and corresponding answer are related to Bangladesh and international affairs. Total of two thousand data is available in this general knowledge dataset. We collect the data from online. Properties of the dataset is given in the table 1.

TABLE 1: Properties of The General Knowledge-Based Question Answering Dataset for Chatbot.

| Dataset properties | Number |
|---|---|
| Question token | 3738 |
| Answer token | 2355 |
| Max input length | 15 |
| Max output length | 10 |
| Total data | 2000 |

*B. Problem Understanding*

The maximum number of sequence-related MT problems are solved by using Encoder and Decoder. That is known as seq2seq learning [1] [2] [3]. Calculating the continuous probability of input text such as word, sentence and phase [4] is main point of this approach.

If we consider the condition probability of our problem is $P(y|x)$. Then $x = x_1,..,x_n$ is the Bengali question which is encoder input. And $y = y_1,..,y_m$ is the answer of the question. Decoder generate the answer using the sequence probability prediction. So, the conditional probability for Bengali question answering chatbot will be,

$$P(y_1,…,y_j|x_1,…,x_i) = \prod_{t=1}^{j} P(y_t|S, y_1,..,y_{t-1}) \quad (1)$$

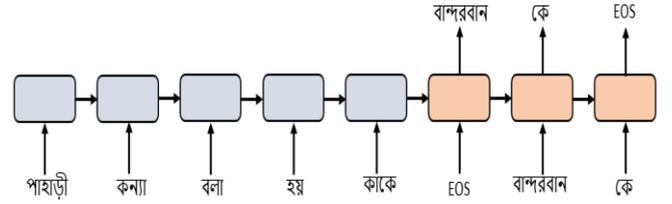

Figure 1: Seq2Seq2 Model for Bengali General Knowledge Question Answering Chatbot.

Generating the sequence and memorized the complex of text sequence LSTM is widely used [5]. Encoder passes the info arrangement with LSTM and Decoder get and gives the yield grouping utilizing decoder LSTM. The yield is produced by the likelihood of decoder with the yield layer the enactment work.

$$P(y_j|S, y_1,..,y_{t-1}) = softmax(W_o. h_{t+1} + b) \quad (2)$$

$y_j$ is the yield of each interpreted expression of the machine. $h_{t+1}$ is a concealed unit with the predisposition b of the initiation work. S is utilized as a source shrouded unit. That keeps up the all-interpretation procedure in the entire model.

*C. Model Architecture*

Multi computational advances and long-haul conditions in successive information are two stages that make a test in any AI framework [9]. For simple QA system recurrence of sequence given a better performance. Transformer is a model that use self-attention model to represent the model input and output without using any recurrence of sequence [6]. In this part we discussed the transformer model that used in making our chatbot.

*a. Encoder*

In the transformer, the encoder has two sub-layers such as multi-head attention and Position wise FFN (feed-forward network). Both sub-layers are connected with a residual connection [10]. Each layer used after layer normalization [11]. Embedding layer is used for the encoder and used 512 dimension of the model and it represent by $d_{model}$.

*b. Decoder*

After adding encoder two sub-layers, decoders added another layer which is known as third sub-layers [6]. That performs multi-head attention. In the decoder layer normalization is also used. Self-attention is also modified in the decoder sub-layers. Masking is used in decode output embedding that assure the prediction position of the output.

*c. Transformer*

The transformer model has encoder and decoder with fully connected layers. Architecture of the transformer model made by the self-attention & position-wise attention mechanism. In figure 2 the left and right cell of the transformer is given.

*i. Scaled Dot-product*

In this attention layer has three parameters such as key, query and value. For calculating the weight of the attention here used dot product of query with all attention keys. And used SoftMax for find the gain of the attention weights. There are two attention such as additive attention [1] and another is dot-product attention [8]. Dot-product attention is faster than any others.

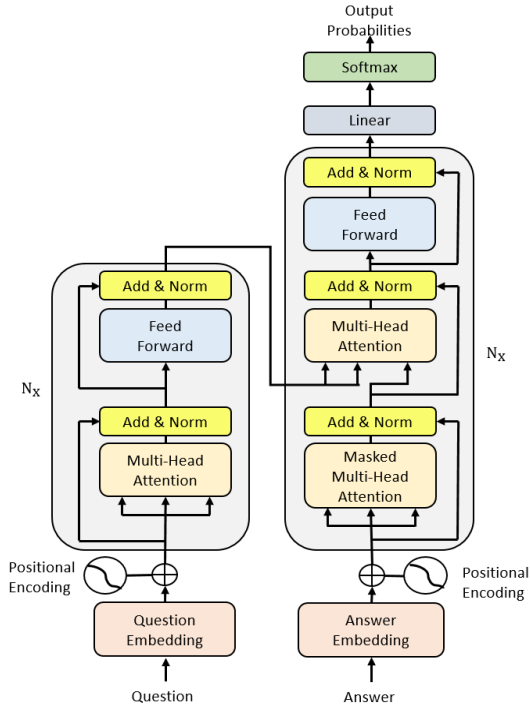

Figure 2: Vaswani [6] transformer model for Bengali Question and Answering chatbot.

*ii. Multi-Head attention*

In attention mechanism dimension is used $d_{model}$ = 512. Three parameters query, value and key are used by linearly projection. Those values are concatenate with head value where head value $h = 8$ that follow parallel attention. After calculating the attention computational cost which is similar with the single-head attention.

*iii. Position-wise FFN*

Transformer encoder and decoder are connected with fully connected FFN. That is used linear function with a activation function. A ReLU function is used in applied activation function. Different parameters are used in each layer.

*iv. Position encoding*

The transformer model has no sequence recurrence. Therefore, it used a position function for maintaining the sequence order. It used relative position of the input embedding sequence. After that the ending of the encoder and decoder the transformer model uses a positional encoding. Each positional encoding has 512 dimensions. Where encoder and decoder are summed up. For calculating the position *sine* and *cosine* function is used. Both functions are calculated the wavelength with the sequence position. Sinusoidal wave length is used which is easily counted the wavelength for the sequence.

## IV. EXPERIMENT RESUT

This section, we discussed the experiment result and performance of the transformer model for our dataset. For the research purpose, we used Seq2Seq model with attention for checking the chatbot performance and compared the result with the transformer model. Both models are performed well for the chatbot. But the transformer gives a more accurate result than Seq2Seq model.

*A. Model Parameters*

The dataset has two parameters one is the question and another is answer text. Parameters of the model are given in table 2. For vocabulary count, we combined both the question and answer. Then the vocabulary is counted for overall text. Total vocabulary size is 2238 for our dataset. We used 2 layers for the transformer model. And set the hidden units 512. 256 hidden units is used in dimension for the model and used 8 head of the model. For reducing the overfitting, we used 0.1 dropout for the transformer model.

TABLE 2: Transformer Model Parameters for Bengali General Knowledge Chatbot.

| Parameters Used for Model | |
|---|---|
| Vocabulary Size | 2238 |
| Number Layers | 2 |
| Units | 512 |
| D_Model | 256 |
| Number Heads | 8 |
| Dropout | 0.1 |

Encoder and decoder are used for making the model. Each encoder and decoder have individual parameters value. Both layers parameters value is given into the table 3.

TABLE 3: Encoder and Decoder Layers Parameter Values.

| Parameters Value for Encoder & Decoder | |
|---|---|
| Vocabulary Size | 2238 |
| Number Layers | 2 |
| Units | 512 |
| D_Model | 128 |
| Number Heads | 4 |
| Dropout | 0.3 |

*B. Training*

We used colab free gpu support for train our model. The dataset contains two thousand questions with the answer. For the training, we used 1741 sample but the rest of the data is used for testing. Batch size of the model is 28 and set the iteration or epoch size is 120. Optimizing our model, we used Adam [7] as optimizer and set value of $\beta_1 = 0.9$ then $\beta_2 = 0.98$, and $\epsilon = 10^{-9}$.

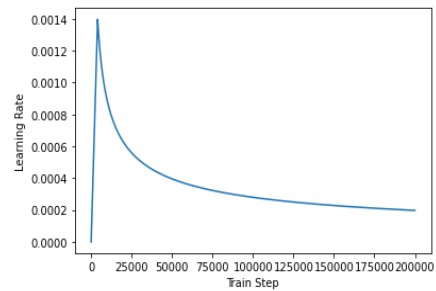

Figure 3: Sample learning rate curve for 20000 train steps.

For calculating the loss function we used sparse categorical as entropy function. We set the learning rate 0.0014 for train the transformer model (Figure3). After training model gave 0.002 loss for trained

*C. Model Evaluation*

We used two approaches for evaluating our chatbot model. At first, we used encoder and decoder based seq2seq model with attention mechanism [8]. The seq2seq model gives 23.5 BLEU for our general knowledge dataset. Then we applied the Transformer for making a better question answering chatbot. Transform gives 85.0 BLEU for chatbot. High BLEU defines the better model for automatic chatbot.

TABLE 4: Applied Models BLEU Score for Dataset.

| Model | BLEU |
|---|---|
| Simple RNN | 12.25 |
| LSTM | 17.13 |
| RNN+GRU | 13.34 |
| Bi-directional RNN | 17.23 |
| Seq2Seq + Global Attention [8] | 23.50 |
| Transformer [6] | **85.00** |

*D. Results*

We set the start and end token for evaluating the model. In the input question we set the both token and set only start token for the output answer. Test the result we split the 20 % of dataset and predict the output. Some tested responses are given in the table 5.

TABLE 5: Transformer Model Response on Test Data.

| Model Response Sample |
|---|
| **Question:** এশিয়ার ক্ষুদ্রতম দেশ কোনটি? |
| **Possible Answer:** মালদ্বীপ |
| *Transformer Response*: মালদ্বীপ |
| **Question:** বাংলাদেশের জাতীয় সংসদের আসন কয়টি? |
| **Possible Answer:** ৩৫০টি |
| *Transformer Response*: ৩৫০টি |
| **Question:** বঙ্গবন্ধু ১ স্যাটেলাইট মহাকাটে অবস্থান করটব ককাথায়? |
| **Possible Answer:** ১১৯.১ ডিগ্রি পূর্ব দ্রাঘিমাংশে |
| *Transformer Response*: ১১৯.১ ডিগ্রি পূর্ব দ্রাঘিমাংশে |

## V. CONCLUSION

An AI-based chatbot presents a more accurate result than any rule-based chatbot application. Also, automatic chatbot application gives an answer on the basis of regarding the question intelligently. Using normal encoder and decoder, seq2seq learning methods perform well for those applications. But sometimes the proper answer remains from the corresponding question. Here we applied the transformer model to find out a better response in a chatbot application. This model performs a better BLEU for the applied Bengali QA dataset. Analyzing the response of the transformer model shows that gives a maximum accurate than seq2seq model. That will help us to make an accurate chatbot system.

ACKNOWLEDGEMENT

Computational Intelligence Lab offers help to achieve our exploration. We give thanks to our Department of Computer Science and Engineering for their support.